\documentclass{article}

\usepackage[preprint]{spconf}

\usepackage{amsmath,graphicx,hyperref}
\usepackage{algorithmic}
\usepackage{array}
\usepackage{makecell}
\usepackage{textcomp}
\usepackage{xcolor}
\usepackage{booktabs}
\usepackage{multirow}
\usepackage{siunitx}
\usepackage{rotating}

\copyrightnotice{\copyright\ IEEE 2026}
\toappear{To appear in {\it Proc.\ ICASSP 2026, May 04-08, 2026, Barcelona, Spain}}

\title{Weighted Temporal Decay Loss for Learning Wearable PPG Data with Sparse Clinical Labels}

\name{%
\begin{tabular}{@{}c@{}}
    \itshape Yunsung Chung$^{1,2}$\sthanks{Work done during an internship at Samsung Research America} \qquad
    Keum San Chun$^{2}$ \qquad
    Migyeong Gwak$^{2}$ \qquad
    Han Feng$^{1}$ \qquad
    Yingshuo Liu$^{1}$ \\
    \itshape Chanho Lim$^{1}$ \qquad
    Viswam Nathan$^{2}$ \qquad
    Nassir Marrouche$^{1}$ \qquad
    Sharanya Arcot Desai$^{2}$
\end{tabular}
}
\address{$^{1}$Tulane University \\ $^{2}$Samsung Research America}
\begin{document}
\maketitle
\begin{abstract}
Advances in wearable computing and AI have increased interest in leveraging PPG for health monitoring over the past decade. One of the biggest challenges in developing health algorithms based on such biosignals is the sparsity of clinical labels, which makes biosignals temporally distant from lab draws less reliable for supervision. To address this problem, we introduce a simple training strategy that learns a biomarker-specific decay of sample weight over the time gap between a segment and its ground truth label and uses this weight in the loss with a regularizer to prevent trivial solutions. On smartwatch PPG from 450 participants across 10 biomarkers, the approach improves over baselines. In the subject-wise setting, the proposed approach averages 0.715 AUPRC, compared to 0.674 for a fine-tuned self-supervised baseline and 0.626 for a feature-based Random Forest. A comparison of four decay families shows that a simple \emph{linear} decay function is most robust on average. Beyond accuracy, the learned decay rates summarize how quickly each biomarker’s PPG evidence becomes stale, providing an interpretable view of temporal sensitivity.
\end{abstract}

%
%

\section{Introduction}
With advances in mobile and wearable computing, it has become ever easier to capture biosignals such as PPG continuously during daily lives. While such advancement leads to proliferation of biosignals, associating them with proper human physiological information (e.g., Cholesterol, HbA1C, Lipids levels, etc.) still remains challenging due to the inherent nature of clinical data collection protocols, which often necessitate blood drawing~\cite{goodwin2022timing}. Since many biomarkers (e.g, lipids) vary over time due to factors such as diet and circadian rhythm, a lab measurement is time-specific, and biosignals collected far from the draw time become less reliable for supervision~\cite{pan2007diurnal}. For this reason, only a small subset of the captured dataset becomes suitable for algorithm development. Given that many advanced AI models need a large amount of data, the reduction in viable, labeled samples presents a substantial roadblock in health algorithm development. 
One naive method to address this problem is to assign a label to all PPG segments within a fixed temporal window. This treats all PPG segments within the window as equally reliable, regardless of their distance to the lab draw. However, this is suboptimal as it introduces substantial noise in labels. Although annotation noise has been extensively studied~\cite{nagaraj2024learning}, methods for leveraging sparse clinical labels to make use of temporally stale biosignals remain underexplored.


To address this challenge, we focus on the temporal distance between a signal and the ground truth label as a proxy for label confidence. We hypothesize that a small time gap indicates a high-confidence sample, where the PPG signal is highly likely to reflect the physiological state at the time of the lab test. In contrast, a large time gap suggests a low-confidence sample with high uncertainty.

In addition, recent studies have demonstrated the potential of PPG for non-invasively estimating various biomarkers. For instance, studies have used PPG to detect potassium imbalances~\cite{miller2023evaluation}, cholesterol levels~\cite{arguello2025non,chen2023non}, creatinine~\cite{sridevi2025integrated}, white blood cells (WBC)~\cite{sabith2025smartphone,barkatullah2024white}, and hemoglobin~\cite{lychagov2023noninvasive}. However, the works typically focus on a single biomarker, often using data from fingertip or clinical-grade devices. Our work expands on this by tackling a diverse panel of 10 biomarkers using noisy, real-world signals from a commercial smartwatch collected in free-living conditions, covering under-explored biomarkers (electrolytes).

\begin{figure*}[h!]
    \centering
    \includegraphics[width=\textwidth]{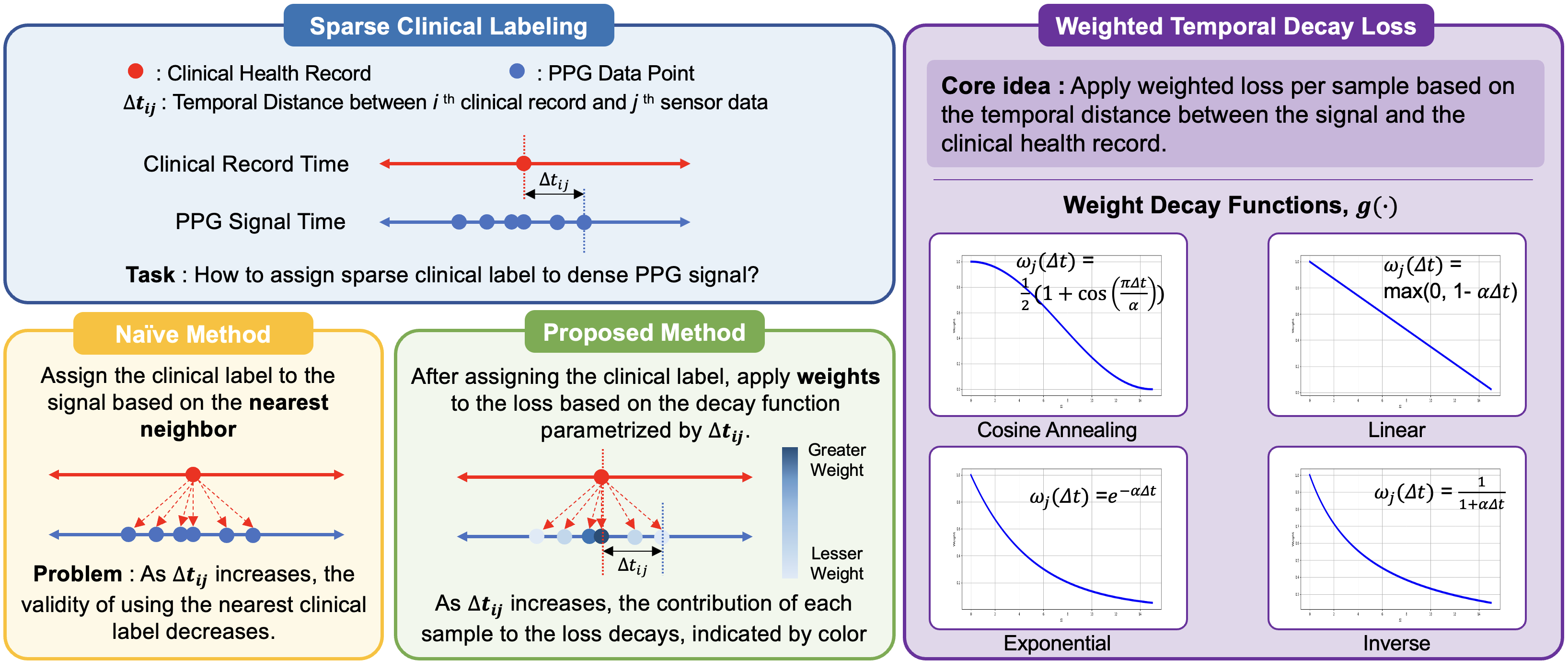}
    \caption{The proposed method uses a weighted decay loss function, parameterized by $\Delta t$ -- the temporal distance between the nearest sparse clinical label and the corresponding sensor data -- to progressively reduce the contribution of the samples as they occur further from the clinical health records.}
    \label{fig:fig1}
  \end{figure*}

In this work, we propose a novel deep-learning framework that directly confronts this issue by learning to leverage all available data, even segments temporally distant from the ground truth labels. Our key contribution is a \textbf{weighted temporal decay loss} that dynamically assigns weights to each sample's contribution to the loss function based on its temporal distance between the biosignal and the ground truth label. By explicitly modeling the temporal gap into training, our model learns a more accurate and generalizable representation between PPG and biomarkers, which improves predictive performance and clinical relevance. 

Our main contributions are:
\begin{itemize}
    \setlength\itemsep{0pt}
    \setlength\parsep{0pt}
    \setlength\topsep{0pt}
    \setlength\partopsep{0pt}
    \item \textbf{Comprehensive Validation on 10 Biomarkers}: We validate that our method improves model performance on 10 biomarkers (lipids, electrolytes, blood counts, A1C) from smartwatch PPG, whereas prior work typically demonstrates results on a single marker and/or clinical-grade sensors.
    \item \textbf{Weighted Temporal Decay Loss}: We introduce a temporally weighted loss function with a learnable decay parameter that improves model robustness and performance by solving the core problem of diminishing relevance between samples and the labels.
    \item \textbf{Data efficiency}: Our method allows deep learning models to be trained on more data by including temporally distant signals to make biomarker research more scalable.
\end{itemize}

\section{Method}

\subsection{Dataset Overview}
A total of 450 participants (236 women, 214 men) collected continuous physiological sensor data between October 2024 and August 2025. The participants ranged in age from 24 to 92 years (M = 59.9, SD = 15.1) and were required to be daily users of a Samsung Galaxy Watch 6. In this analysis, we use green PPG captured at a sampling rate of 25Hz. The signals are labeled with clinical health records encompassing 10 biomarkers shown in Table \ref{tab:scenario2}. 
Raw PPG signals were segmented into non-overlapping 10\,s segments. Then, a signal-quality index (SQI) was assigned to each segment to discard low-quality data. For the resulting high SQI segments, we applied a 0.5--5Hz band-pass filter and z-score normalization. The resulting PPG segments were assigned a ground truth label based on the nearest clinical health record. If the nearest clinical record was greater than 30 days, the segment was not included in the analysis.

\subsection{Weighted Temporal Decay Loss}
Despite the abundance of biosignals, sparse clinical labels remain a major bottleneck for data-hungry deep learning models in health applications~\cite{miotto2018deep}. A naive approach of matching each lab result with the temporally closest PPG segment restricts the dataset size (See Figure \ref{fig:fig1}). However, it is suboptimal as it does not account for the diminishing relevance of the clinical records and the corresponding sensor data. This leads to a problem where the model is forced to treat the temporally distant signals with equal confidence~\cite{nagaraj2024learning,miotto2018deep}. Our approach addresses this by introducing a weighted temporal decay loss designed to account for the diminishing relevance.

In our proposed method, rather than treating all samples equally regardless of whether proximal or distal from the ground truth label, we assign a weight to each PPG segment based on the temporal distance between the sample and the clinical ground truth label (Figure \ref{fig:fig1}). For a processed 10-second PPG segment at $i$-th index, let $\Delta t_i\!\ge\!0$ be the absolute temporal distance measured in \emph{days} between the segment's median timestamp and the nearest health record time (i.e., the time at which the health record was obtained). Then, we define the weight $w_i$ for the corresponding $i$-th segment as the following:
\[
w_i = g\!\big(\hat\alpha_b\,\Delta t_i\big), \qquad \hat\alpha_b=\mathrm{softplus}(\alpha_b)\ge 0,
\]
where $g(\cdot)$ is a monotonically decreasing decay function, and $\hat\alpha_b$ is a learnable decay rate. The $\hat\alpha_b$ defines the rate at which the weight decays as $\Delta t_i$ increases, and it is independently learned for each biomarker. The weight $w_i$ is used to represent the diminishing relevance between the sensor data and the ground truth health record. For example, as $\Delta t_i$ increases, the sensor data becomes less relevant to the corresponding ground truth health record. Thus, the contribution of the corresponding sample to the overall loss is reduced, as parameterized by $w_i$.

We train with a weighted binary cross-entropy (BCE) plus a simple mean-weight bonus that discourages the degenerate solution $w_i\!\approx\!0$ (i.e., excessively large $\hat\alpha_b$). The weight loss employed in this study is expressed as the following:
\begin{align}
\mathcal{L}_{\text{weighted}}
= \frac{1}{N}\sum_{i=1}^{N} w_i\,\mathrm{BCE}(\hat y_i,y_i)
\;-\;\lambda\,\underbrace{\frac{1}{N}\sum_{i=1}^{N} w_i}_{\bar w}
\end{align}
where $N$ is the number of segments, $\mathcal{L}_{\text{weighted}}$ is the weighted loss, $y_i$ is the ground truth label, $\hat y_i$ is the predicted label and $\lambda$ is the regularization weight. This keeps training focused on closer segments while still leveraging useful distant ones; gradients flow through $w_i$ and the softplus to update $\alpha_b$ to learn the biomarker-specific temporal sensitivity.

We do not normalize $\Delta t_i$ by the window, and we do not apply $w_i$ at inference time. We apply $w_i$ \emph{only} during training. Inference uses the base network without weighting, so there is no runtime or deployment overhead. Unless otherwise noted, all time distances are measured in \emph{days}, and the window length (30 days) bounds the support of $\Delta t_i$. To avoid overfitting, we fixed $\lambda=0.5$ across all experiments and did not tune it per biomarker or fold. 

In this study, we explored four different decay functions -- exponential, linear, cosine annealing, and inverse -- shown in Figure \ref{fig:fig1}, and among the four decay families, the \emph{Linear} schedule proved most robust on average (Table~\ref{tab:decay_comparison}). Thus, we adopt it for the primary comparisons.

\subsection{Experimental Setup}
In this analysis, we include only the two extreme quantiles (i.e., top 25\% and bottom 25\%) of the clinical recordings for each biomarker to examine the feasibility of distinguishing the two quantiles strictly using PPG morphology. To mitigate subject-wise imbalance, we capped the number of segments contributed by subject per biomarker at the median across all subjects. The upper quantile (i.e., top 25\%) was assigned a positive label, and the bottom quantile (i.e., bottom 25\%) was assigned a negative label. The final count of subjects per biomarker after SQI filtering and windowing is summarized in Table \ref{tab:scenario2}. 

We use 5-fold cross-validation stratified by subject to ensure subject independence between folds. We report the model performance in AUROC and AUPRC. For comparison, we use two baseline models: a feature-based Random Forest and a state-of-the-art self-supervised learning method, PAPAGEI~\cite{pillai2024papagei} with fine-tuning. The random forest model was trained on a set of 34 handcrafted features, including 28 morphology descriptors (e.g., pulse time, augmentation index) and 5 time-domain heart rate variability (HRV) metrics (e.g., SDNN, RMSSD), and the average heart rate (HR). The PAPAGEI model was used as it was trained on a large dataset and pretrained on PPG morphological features~\cite{pillai2024papagei}.

We use the same PAPAGEI pretrained encoder and classifier head as PAPAGEI (FT), changing the training objective to the proposed weighted temporal decay loss. We report a two-extreme protocol (top vs. bottom quartile) to test feasibility under sparse, temporally misaligned labels. Performance in the middle 50\% is not assessed, and full-distribution modeling is left for future work.

\section{Results}

\begin{table}[tb]
\centering
\caption{Model performance comparison. AUROC/AUPRC are reported. Predictions are formed by averaging segment \emph{logits} per subject. 
$N$ denotes unique subjects (and the number of valid segments) with at least one eligible segment in the 30-day window.}
\label{tab:scenario2}
\resizebox{\columnwidth}{!}{%
\begin{tabular}{l r c c c}
\toprule
\textbf{Biomarker} & \textbf{N} & \textbf{RF} & \textbf{PAPAGEI (FT)} & \textbf{Ours} \\
\midrule
LDL           & 39 (7,521) & 0.465/0.477 & 0.526/0.473 & \textbf{0.644/0.507} \\
Triglyceride  & 33 (7,880) & 0.547/0.612 & \textbf{0.740/0.877} & 0.715/0.840 \\
HbA1C           & 42 (7,719) & 0.660/0.603 & 0.703/0.637 & \textbf{0.727/0.657} \\
Hemoglobin    & 69 (16,643) & 0.680/0.645 & 0.797/0.804 & \textbf{0.798/0.815} \\
CO\textsubscript{2}    & 77 (20,992) & 0.546/0.694 & 0.536/0.667 & \textbf{0.650/0.720} \\
Chloride      & 86 (20,310) & 0.494/0.571 & 0.703/0.725 & \textbf{0.716/0.750} \\
Potassium     & 72 (16,847) & 0.599/0.607 & 0.676/0.612 & \textbf{0.724/0.713} \\
Sodium        & 73 (16,417) & 0.670/0.635 & 0.522/0.518 & \textbf{0.593/0.604} \\
WBC           & 57 (13,895) & 0.736/0.795 & 0.755/0.745 & \textbf{0.843/0.848} \\
Platelets     & 56 (15,460) & 0.600/0.632 & 0.660/0.699 & \textbf{0.713/0.698} \\
\midrule
\textbf{Average} & -- & 0.599/0.626 & 0.660/0.674 & \textbf{0.712/0.715} \\
\bottomrule
\end{tabular}}
\end{table}

\subsection{Two-Extreme Classification}
We aggregate all eligible segments per individual into a single prediction by averaging segment logits within the window, yielding one decision per person (Table~\ref{tab:scenario2}). Our model averages 0.712 AUROC / 0.715 AUPRC, outperforming PAPAGEI (0.660 / 0.674) and RF (0.599 / 0.626). The effect is visible on markers with faster physiology: Potassium reaches 0.724 AUROC and WBC 0.843 AUROC. Aggregation reduces momentary noise, and the weighted temporal decay loss tempers the influence of segments that are too far from the lab time, which together yield a clearer summary per patient. On average, this corresponds to an improvement of +0.052(+7.8\%) AUROC and +0.041(6.1\%) AUPRC over PAPAGEI.

\subsection{Temporal Decay Functions}
We examined four decay families under the identical conditions (e.g., the same data splits, preprocessing, initialization, and training hyperparameters). The only difference was the weighting function. As shown in Table~\ref{tab:decay_comparison}, a \emph{Linear} decay function gives the strongest average (0.712 / 0.715), with Inverse and Exponential close behind, and Cosine Annealing. We hypothesize that Linear is robust here for two reasons: (i) it provides a cut-off truncation at $\Delta t \approx 1/\alpha$, eliminating gradients from very distant segments, and (ii) it downweights steadily without the early saturation of Exponential or the long tails of Inverse. The best decay function is expected to be biomarker- and dataset-dependent (e.g., distribution of time gaps, within-subject variability). On our dataset, Linear showed the strongest average, but per-biomarker results varied. Thus, decay selection should consider the temporal characteristics of the target biomarker and the empirical distribution of $\Delta t$. 

\subsection{Ablations}
Ablation results indicate two key design choices. First, removing the time-aware loss drops the average to 0.660 / 0.674 (Table~\ref{tab:ablation_study}), roughly the level of the PAPAGEI baseline. Second, keeping the loss but fixing the decay rate yields 0.676 / 0.694. Learning a biomarker-specific rate adds another +0.036 AUROC and +0.021 AUPRC over that fixed version. In short, temporal weighting by $\Delta t$ carries most of the benefit, and learning the rate tightens it further.

\begin{table}[tb]
\centering
\caption{Decay families under the subject-wise protocol.}
\label{tab:decay_comparison}
\begin{tabular}{lcc}
\toprule
\textbf{Decay} & \textbf{AUROC} & \textbf{AUPRC} \\
\midrule
Linear       & \textbf{0.712} & \textbf{0.715} \\
Inverse      & 0.691 & 0.702 \\
Exponential  & 0.689 & 0.687 \\
Cosine Annealing      & 0.668 & 0.690 \\
\bottomrule
\end{tabular}
\end{table}

\begin{table}[tb]
\centering
\caption{Ablation study results for our time-aware model. Performance is reported as average AUROC / AUPRC}
\label{tab:ablation_study}
\begin{tabular}{@{}lcc@{}}
\toprule

\textbf{Model Configuration} & AUROC & AUPRC \\
\midrule
Our Full Model  & \textbf{0.712} & \textbf{0.715} \\
\quad -- w/o Learnable $\alpha$ & 0.676 & 0.694 \\ 
\quad -- w/o Time-aware Loss & 0.660 & 0.674 \\
\bottomrule
\end{tabular}
\end{table}

\section{Discussion}
Overall, the time-aware model leads across most biomarkers in the experiment. A notable exception is \emph{Triglyceride}, where PAPAGEI outperforms the proposed approach (0.740/0.877 vs.\ 0.715/0.840). Two factors may explain this: (i) the triglycerides cohort is the smallest in our study ($N=33$ subjects), which increases variance, and (ii) triglycerides are strongly influenced by recent diet, so a single lab draw may poorly anchor nearby segments in the 30-day window. In contrast, markers with faster dynamics (e.g., WBC) are more sensitive to temporal misalignment and thus gain more from the weighted temporal decay loss~\cite{he2018circadian}.


\noindent\textbf{Limitation.} Our experiments use a fixed 30-day window, but different biomarkers would likely need different horizons for optimal performance. A natural extension is to learn the window length (or a soft truncation) jointly with the decay rate, allowing the model to choose how far back to look for each biomarker. This can be done by parameterizing a differentiable mask over time gaps, which would preserve end-to-end training while avoiding hand-tuned windows.

Our study uses data from a single device family and a single health system, so domain shift (e.g., different hardware, demographics, or lab workflows) may affect performance; external validation remains important. The window length was fixed at 30 days for the main results, while ablations suggest robustness, very long windows can change label prevalence, and should be tuned to the application. The study was reviewed under institutional oversight with informed consent and de-identified data. 

\section{Conclusion}
In this work, we addressed the challenge of diminishing relevance in labeled PPG signal due to the sparse ground truth labeling in predicting 10 biomarker levels. We introduced a novel time-aware model that incorporates a decay function to dynamically weight signal information based on its temporal proximity to the corresponding lab test.
Our experimental results demonstrate that incorporating time-gap–aware weighting during training is an effective strategy for non-invasive biomarker prediction under sparse clinical labels. Across 10 biomarkers, we observe consistent improvements over strong baselines, including electrolytes and lipids that are challenging for non-invasive sensing. Our method represents a step forward in developing a practical, continuous, and non-invasive patient monitoring system.
Future work should focus on validating this model on external datasets from diverse clinical environments to ensure generalizability. 

\bibliographystyle{IEEEbib}
\bibliography{strings,refs}

\end{document}